\documentclass[10pt,twocolumn,letterpaper]{article}

\usepackage[pagenumbers]{iccv}              

\definecolor{iccvblue}{rgb}{0.21,0.49,0.74}

\usepackage{amssymb}
\usepackage{amsmath}
\usepackage{pifont}
\usepackage{bm}
\usepackage{booktabs}
\usepackage{mathtools}
\usepackage{multirow}

\usepackage[numbers]{natbib}
\usepackage[resetlabels]{multibib}
\newcites{supp}{Supplementary References}
\usepackage[pagebackref,breaklinks,colorlinks,citecolor=iccvblue,allcolors=iccvblue]{hyperref}
\usepackage{etoolbox}
\makeatletter
\patchcmd\Hy@backout{\@auxout}{\@mainaux}{}{\fail}
\patchcmd\Hy@backout{\@auxout}{\@mainaux}{}{\fail}
\makeatother

\newcommand{\myparagraph}[1]{\vspace{2mm}\noindent\textbf{#1}}

\usepackage{balance}
\usepackage{xcolor}
\usepackage{color,colortbl}

\definecolor{col_nose}{RGB}{181,228,140}
\definecolor{col_mouth}{RGB}{255,183,3}
\definecolor{col_forehead}{RGB}{189,178,255}
\definecolor{col_cheek}{RGB}{144,224,239}
\definecolor{col_table}{RGB}{175,227,246}

\definecolor{mycolor1}{rgb}{0.85000,0.32500,0.09800}
\definecolor{mycolor2}{rgb}{0.92900,0.69400,0.12500}
\definecolor{mycolor3}{rgb}{0.49400,0.18400,0.55600}
\definecolor{mycolor4}{rgb}{0.87843,0.76471,0.98824}
\definecolor{mycolor5}{rgb}{0.46600,0.67400,0.18800}
\definecolor{mycolor6}{rgb}{0.30100,0.74500,0.93300}
\definecolor{mycolor7}{rgb}{0.00000,0.44700,0.74100}

\definecolor{best_two}{RGB}{72,149,239}
\definecolor{best}{RGB}{179,11,0}

\usepackage[lined,ruled,linesnumbered]{algorithm2e}

\title{SRM-Hair: Single Image Head Mesh Reconstruction via 3D Morphable Hair}

\author{Zidu Wang$^{1,2}$, Jiankuo Zhao$^{1,2}$, Miao Xu$^{1,3}$, Xiangyu Zhu$^{1,2}$\footnotemark[1], Zhen Lei$^{1,2,3}$\\
    $^{1}$State Key Laboratory of Multimodal Artificial Intelligence Systems,\\ Institute of Automation, Chinese Academy of Sciences\\
    $^{2}$School of Artificial Intelligence, University of Chinese Academy of Sciences\\
    $^{3}$ Centre for Artificial Intelligence and Robotics, Hong Kong Institute of Science \& Innovation,\\ Chinese Academy of Sciences\\
    {{\tt\small wangzidu0705@gmail.com}, \tt\small \{zhaojiankuo2024, zhen.lei\}@ia.ac.cn}, {\tt\small miao.xu@cair-cas.org.hk},\\{\tt\small xiangyu.zhu@nlpr.ia.ac.cn}
}

\begin{document}

\twocolumn[{
\renewcommand\twocolumn[1][]{#1}
\maketitle
\begin{center}
\includegraphics[width=\textwidth]{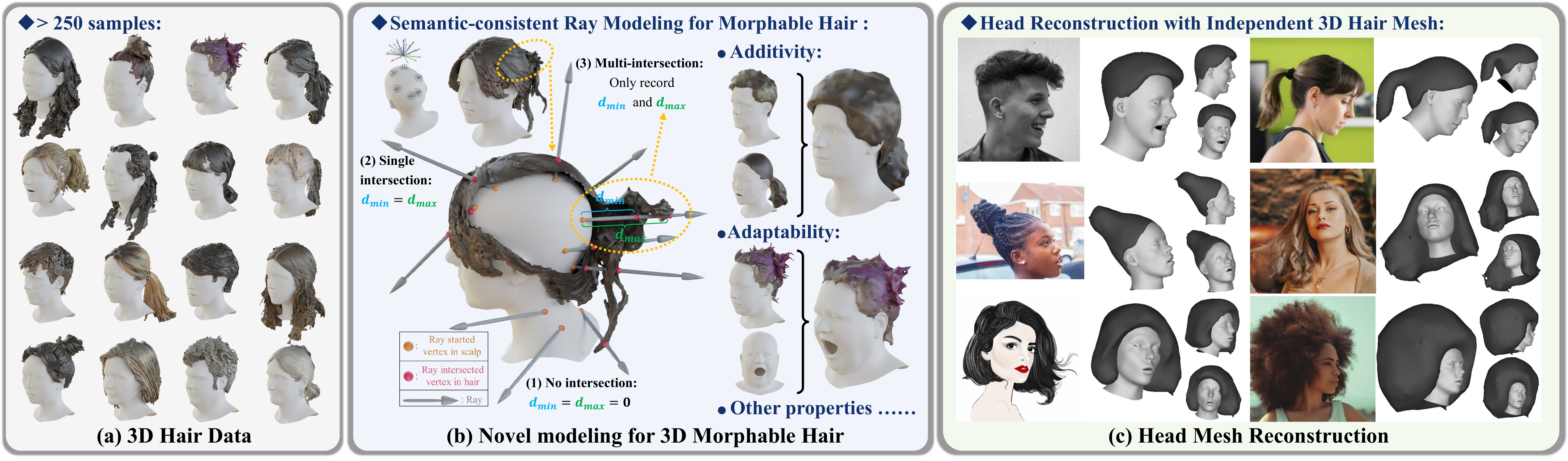}
\vspace{-0.6cm}
\captionof{figure}{We propose SRM-Hair to advance head mesh reconstruction with a dataset of over 250 high-fidelity real hair mesh scans paired with 3D face data (a), along with a novel method for reconstructing a 3D head mesh with independent hair region from a single image (c) using semantic-consistent ray modeling (b) for 3D morphable hair.
}
\label{teaser}
\end{center}
\vspace{-0.3cm}
}]

{\renewcommand{\thefootnote}{\fnsymbol{footnote}}
    \footnotetext[1]{Corresponding author.}}

\begin{abstract}{

3D Morphable Models (3DMMs) have played a pivotal role as a fundamental representation or initialization for 3D avatar animation and reconstruction. However, extending 3DMMs to hair remains challenging due to the difficulty of enforcing vertex-level consistent semantic meaning across hair shapes. This paper introduces a novel method, Semantic-consistent Ray Modeling of Hair (SRM-Hair), for making 3D hair morphable and controlled by coefficients. The key contribution lies in semantic-consistent ray modeling, which extracts ordered hair surface vertices and exhibits notable properties such as additivity for hairstyle fusion, adaptability, flipping, and thickness modification. We collect a dataset of over 250 high-fidelity real hair scans paired with 3D face data to serve as a prior for the 3D morphable hair. Based on this, SRM-Hair can reconstruct a hair mesh combined with a 3D head from a single image. Note that SRM-Hair produces an independent hair mesh, facilitating applications in virtual avatar creation, realistic animation, and high-fidelity hair rendering. Both quantitative and qualitative experiments demonstrate that SRM-Hair achieves state-of-the-art performance in 3D mesh reconstruction. Our project is available at \href{https://github.com/wang-zidu/SRM-Hair}{https://github.com/wang-zidu/SRM-Hair}.

}
\end{abstract}

\section{Introduction}

3DMMs \cite{blanz1999morphable, blanz2003face, FLAME:SiggraphAsia2017, egger20203d} have significantly shaped the 3D reconstruction and animation of human-related objects, yet they fail to represent the hair region, even though hair meshes are widely recognized as a powerful representation in various applications \cite{bhokare2024real, yuksel2009hair,liu2025lucas}. On the one hand, in high-fidelity hair and head rendering, the reconstruction of hair mesh geometry is a critical guidance and foundation for strand-based hair modeling \cite{chai2016autohair,hu2015single,wu2024monohair,Wu_2022_CVPR}. Head reconstruction with hair mesh is also vital for downstream tasks such as 3D avatars \cite{vhap,qian2024gaussianavatars,wang2024mega}. On the other hand, in computationally constrained environments, precise hair mesh geometry, combined with appropriate textures or synthesis techniques, can yield visually compelling results \cite{bhokare2024real,yuksel2009hair,unihair}. Studies have shown that hair meshes can be effectively employed to animate hair motion \cite{wu2016real}.

This paper focuses on scenarios where a 3D head with hair mesh is required from in-the-wild input. The development of hair mesh modeling and reconstruction has encountered significant challenges, primarily due to the lack of suitable data and effective modeling techniques. Regarding data, although handcrafted datasets or scanning techniques \cite{shen2023CT2Hair, giebenhain2023nphm, wang2022faceverse} offer high geometric accuracy, they necessitate specialized equipment or skilled artists. Moreover, these methods often produce hair meshes entangled with irrelevant elements such as clothing \cite{giebenhain2023nphm, wang2022faceverse} or lack corresponding real 3D face data \cite{shen2023CT2Hair}. Generation-based reconstruction approaches \cite{TRELLIS, long2023wonder3d, liu2023one2345, liu2023one2345++} are capable of generating 3D heads but suffer from inaccuracies caused by issues like the Janus problem \cite{babiloni2024id}. NeRF-based methods \cite{mildenhall2020nerf, Feng2023DELTA, mueller2022instant} typically require a large number of multi-view images as input, making them computationally expensive and time-consuming. Additionally, these methods often necessitate reprocessing or retraining when dealing with new identities. Regarding hair modeling, unlike the well-known 3DMMs \cite{blanz1999morphable, blanz2003face,FLAME:SiggraphAsia2017} for 3D faces, the lack of dedicated methods for establishing point-to-point correspondence in hair surfaces hinders the reconstruction of 3D hair regions, preventing it from being as convenient as single-image-based 3D face reconstruction \cite{guo2020towards,deng2019accurate,DECA:Siggraph2021,wang20243d,wang2024s2td,10506678,10127617}. Some methods \cite{wu2024monohair,vhap,Feng2023DELTA} rely on multi-view images or video inputs to model hair mesh, which introduces limitations such as extensive data preprocessing and substantial computational overhead.

We proposes a novel \textbf{S}emantic-consistent \textbf{R}ay \textbf{M}odeling for reconstructing a 3D head with independent \textbf{Hair} mesh from a single image (\textbf{SRM-Hair}) through 3D morphable hair. As illustrated in Fig.~\ref{teaser}(a), we organized experienced researchers to extract 3D hair from high-fidelity 3D head scans \cite{wang2022faceverse, giebenhain2023nphm}, resulting in a dataset of over 250 accurate hair mesh scans independent of irrelevant regions such as the face and clothing. Notably, our dataset is entirely derived from real scans and includes paired 3D face data along with the original full-head scans, facilitating seamless integration into avatar design and benefiting communities. To build 3D morphable hair based on our data, SRM-Hair emphasizes the importance of obtaining an ordered sequence of 3D vertices on the hair surface. To achieve this, SRM-Hair defines a fixed multi-ray direction template and symmetrically places these ray sets on the scalp surface. Our modeling method offers several outstanding properties, as illustrated in Fig.~\ref{teaser}(b) and Fig.~\ref{properties}, the additivity allows the fusion of multiple hairstyle patterns into a single sample, while adaptability enables transferring known hair data to a head with new identity and expression. Furthermore, it supports flipping hairstyles and modifying hair thickness. To the best of our knowledge, it is the first hair modeling method that allows the representation of a new hairstyle through a combination of known 3D hair samples controlled by coefficients, similar to 3DMMs \cite{blanz1999morphable, blanz2003face, wood2021fake}.

Based on high-quality data and morphable hair with the novel modeling approach, SRM-Hair introduces a novel method for reconstructing a 3D hair mesh from a single image. During the reconstruction process, SRM-Hair predicts the coefficients of the 3D morphable hair corresponding to the input image to generate a hair mesh. The result is refined by a specialized backbone that removes invalid points, resulting in the final reconstruction, as illustrated in Fig.~\ref{teaser}(c). The training process, combining 3D data and 2D in-the-wild images, equips the method with the ability to generalize to real-world scenarios. SRM-Hair makes hair mesh reconstruction as convenient and efficient as monocular 3D face reconstruction while achieving state-of-the-art performance in extensive quantitative and qualitative experiments. Our main contributions are as follows: 

\begin{itemize}
\item{\textbf{Data:} We annotate a dataset of over 250 accurate hair mesh scans, including paired 3D face data and the original full-head scans.}

\item{\textbf{Modeling:} We introduce a novel semantic-consistent ray modeling approach for hair meshes, enabling the extraction of hair surface vertices in a specific order. The modeling yields properties such as additivity, adaptability, {\etc}, resulting in a 3D morphable hair.}

\item{\textbf{Reconstruction:} We propose SRM-Hair, which reconstructs a 3D hair mesh from a single image via 3D morphable hair. Experiments show that SRM-Hair achieves excellent performance and surpasses existing methods.}

\end{itemize}

\begin{figure*}[t]
\begin{center}
\includegraphics[width=1\linewidth]{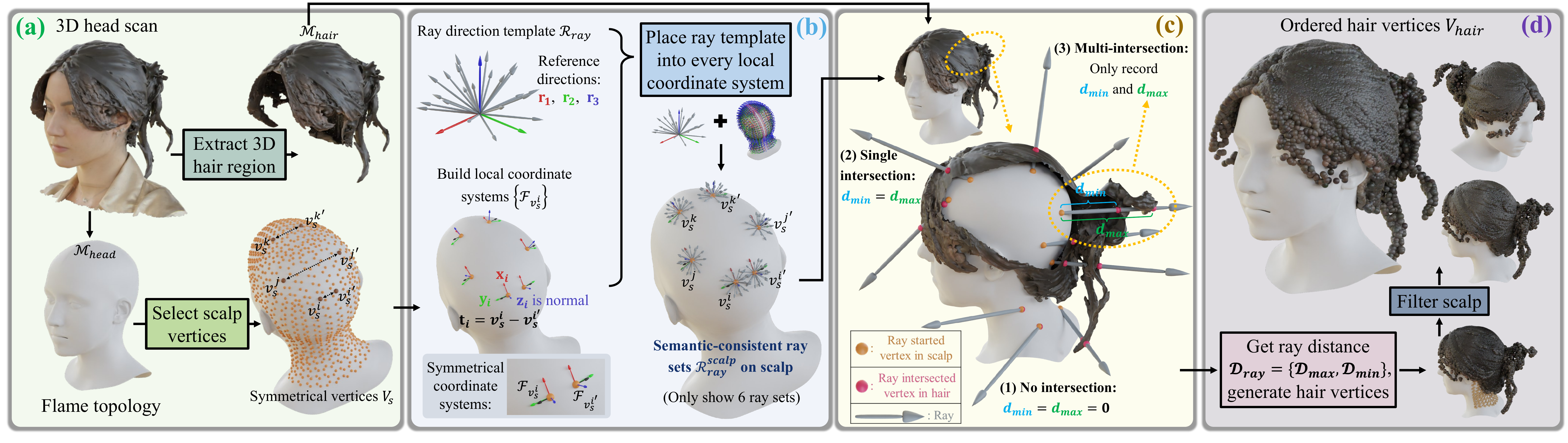 }
\end{center}
\vspace{-0.4cm}
\caption{Overview of our semantic-consistent ray modeling. The core idea is to emit sufficient semantically consistent rays with specific origins and directions to intersect with the hair surface, enabling statistical analysis of the hair vertices.
   }
\label{modeling}
\vspace{-0.4cm}
\end{figure*}

\section{Related Work}

\myparagraph{3D Hair Data.} 3D hair data suffers from limited sample availability and a shortage of real-world examples. Existing 3D hair strand datasets \cite{chai2016autohair,hu2015single,shen2023CT2Hair} typically rely on manual annotation or specialized equipment ({\eg}, CT scanners), which raises the usage barrier, limits samples, and incurs high computational costs. Some studies provide hair mesh data \cite{bhokare2024real,yuksel2009hair,unihair} but often lack correspondence with real heads. Extracting 3D hair meshes from full-head scans \cite{giebenhain2023nphm,wang2022faceverse,ramon2021h3d} appears to be a better alternative, as 3D full-head scans are more abundant and come with matching real 3D faces, clothing, and other assets.

\myparagraph{Hair Modeling and Reconstruction.} Strand-based hair modeling \cite{wu2024monohair, Wu_2022_CVPR, chai2016autohair, hu2015single, sklyarova2023neural} can produce high-fidelity hair renderings but comes with significant computational costs. Some studies explore more efficient hybrid representations for 3D hair modeling. NeRF-based \cite{raina2025prismavatar, Feng2023DELTA} and 3DGS-based \cite{luo2024gaussianhair, zakharov2024gh, unihair} methods have emerged, but they often require video input or are restricted to specific viewpoints, limiting their practicality. Other works attempt to model hair meshes \cite{sevastopolsky2023headcraft, giebenhain2024mononphm,zhan2024shert,caselles2023implicit,ma2021pixel}, but they do not separate hair from the face and struggle to capture complex hairstyles, such as afro-hair or ponytails. Generation-based methods \cite{babiloni2024id, TRELLIS, long2023wonder3d, liu2023one2345, liu2023one2345++, an2023panohead} can reconstruct head mesh from a single image but suffer from issues such as the Janus problem. Some works \cite{zhang2025disentangled, liu2025lucas} attempt to disentangle hair, yet accurate hair geometry estimation and modeling remain crucial.

\section{Modeling}
This section introduces the 3D hair mesh dataset we provide and the modeling method employed by SRM-Hair. Specifically, SRM-Hair utilizes semantic-consistent rays to obtain an ordered sequence of 3D vertices on the hair surface. The method for reconstructing a 3D hair mesh from a single image will be detailed in Sec.~\ref{4Reconstruction}.

\subsection{Data Preparation}

As shown in Fig.~\ref{modeling}(a), given a high-fidelity 3D full-head scan, we collaborate with experienced researchers to extract the 3D hair mesh. The data is constructed from two sources: FaceVerse \cite{wang2022faceverse} and NPHM \cite{giebenhain2023nphm}. Based on the data, we carefully select over 250 representative samples and annotate their hair meshes (if horizontally flipped, there are more than 500 samples). NPHM provides FLAME-registered meshes corresponding to the scans, while for FaceVerse, we obtain head meshes with FLAME topology based on \cite{FLAME:SiggraphAsia2017,FLAME_fitting_github}. Results are shown in Fig.~\ref{teaser}(a). Note that our hair meshes are entirely derived from real scans and include aligned 3D face meshes along with the corresponding full-head scans, allowing for integration into avatar design and providing a valuable asset for the community.

Given an aligned pair of FLAME head mesh ${{\cal M}}_{head}$ and hair mesh ${{\cal M}}_{hair}$, SRM-Hair selects $N_s$ semantically symmetric vertices $V_s = \{v_s^i\}_{i=1}^{N_s} \in {{\mathbb{R}}^{3 \cdot  {N_s}}}$ from the scalp region of ${{\cal M}}_{head}$. As shown in the lower right part of Fig.~\ref{modeling}(a), for $\forall i \in \{1, \dots, N_s\}$, there exists a corresponding vertex $v_s^{i'} \in V_s$ that is symmetric to $v_s^i$ on the scalp surface. Benefiting from the consistent topology of the FLAME model, the semantic position and the symmetry of $v_s^i$ remain invariant by variations in pose, expression, and identity. To simplify the notation, we divide $V_s$ into two parts, {\ie}, $V_s = V_s^{{left}} \cup V_s^{{right}}$.

\subsection{Semantic-consistent Ray Modeling for Hair}

The core idea of our modeling method is to emit a sufficient number of semantically consistent rays with specific origins and directions to obtain the vertices where they intersect with the hair surface, allowing for statistical analysis of the hair vertices. 

\myparagraph{Get Multi-ray Direction Template.} As shown in Fig.~\ref{modeling}(b), a multi-ray direction template $\mathcal{R}_{ray} = \{\mathbf{r}_n\}_{n=1}^{N_r}$ is pre-defined, with the directions sampled from the subdivisions of a hemisphere. Once $\mathcal{R}_{{ray}}$ is defined, it remains fixed and contains three reference directions $\mathbf{r}_1$, $\mathbf{r}_2$, and $\mathbf{r}_3$, which are mutually perpendicular vectors forming a right-handed coordinate system, satisfying $\mathbf{r}_1 \times \mathbf{r}_2 = \mathbf{r}_3$.

\myparagraph{Build Symmetrical Local Coordinate Systems.} The goal of this process is to place $\mathcal{R}_{{ray}}$ at each scalp vertex $v_s^i \in V_s$. We aim for the construct of semantic-symmetrical local coordinate systems. Specifically, for $\forall v_s^{i} \in V_s$, we select the normal direction at $v_s^{i}$ on the head surface as the z-axis ${\bf{z}}_i$ and compute the direction $\mathbf{t}_i = \bm{v}_s^{i} - \bm{v}_s^{i'}$ to construct the local coordinate system ${\mathcal{F}_{v_s^i}} = \langle {{\mathbf{x}}_i},{{\mathbf{y}}_i},{{\mathbf{z}}_i}\rangle$:
\begin{equation}
\left\{
\begin{aligned}
\mathbf{x}_i'&= \mathbf{t}_i \times \mathbf{z}_i \\
\mathbf{y}_i &= \mathbf{z}_i \times \mathbf{x}_i' \\
\mathbf{x}_i &= \delta  \cdot   \mathbf{x}_i' 
\end{aligned}\text{   ,}
\right.
\end{equation}
where if $v_s^{i} \in V_s^{{left}}$, $\delta = +1$, otherwise $\delta = -1$. As shown in Fig.~\ref{modeling}(b), this setup ensures that $\mathcal{F}_{v_s^i}$ is a right-handed coordinate system, while its semantically symmetric system $\mathcal{F}_{v_s^{i'}}$ is a left-handed. This setting preserves the semantic symmetry of local coordinate systems, which is overlooked in \cite{qian2024gaussianavatars, rong2024gaussiangarments}. These works \cite{qian2024gaussianavatars, rong2024gaussiangarments} select one triangle edge as the x-axis, disregarding the symmetry.

\myparagraph{Place Template $\mathcal{R}_{ray}$ into $ \mathcal{F}_{v_s^i}$.} As shown in the right of Fig.~\ref{modeling}(b), we replicate the fixed ray direction template $\mathcal{R}_{{ray}}$ $N_s$ times and place them in each local coordinate system $\mathcal{F}_{v_s^i}$, such that the three reference directions of the template, $\mathbf{r}_1$, $\mathbf{r}_2$, and $\mathbf{r}_3$, align with the three coordinate axes $\mathbf{x}_i$, $\mathbf{y}_i$, and $\mathbf{z}_i$ of $\mathcal{F}_{v_s^i}$. Specifically, when handling the case for $V_s^{{right}}$, we first flip $\mathcal{R}_{{ray}}$ into a left-handed template before placing it.

Consequently, we construct the semantic-consistent ray sets on the scalp $\mathcal{R}_{{ray}}^{scalp} = \{\mathcal{R}_{{ray}}^i\}_{i=1}^{N_s}$, where the order, origin, direction, and symmetry properties of each ray are semantic-consistent and independent of the head's identity and expression variations.

\begin{figure}[t]
\begin{center}
\includegraphics[width=1\linewidth]{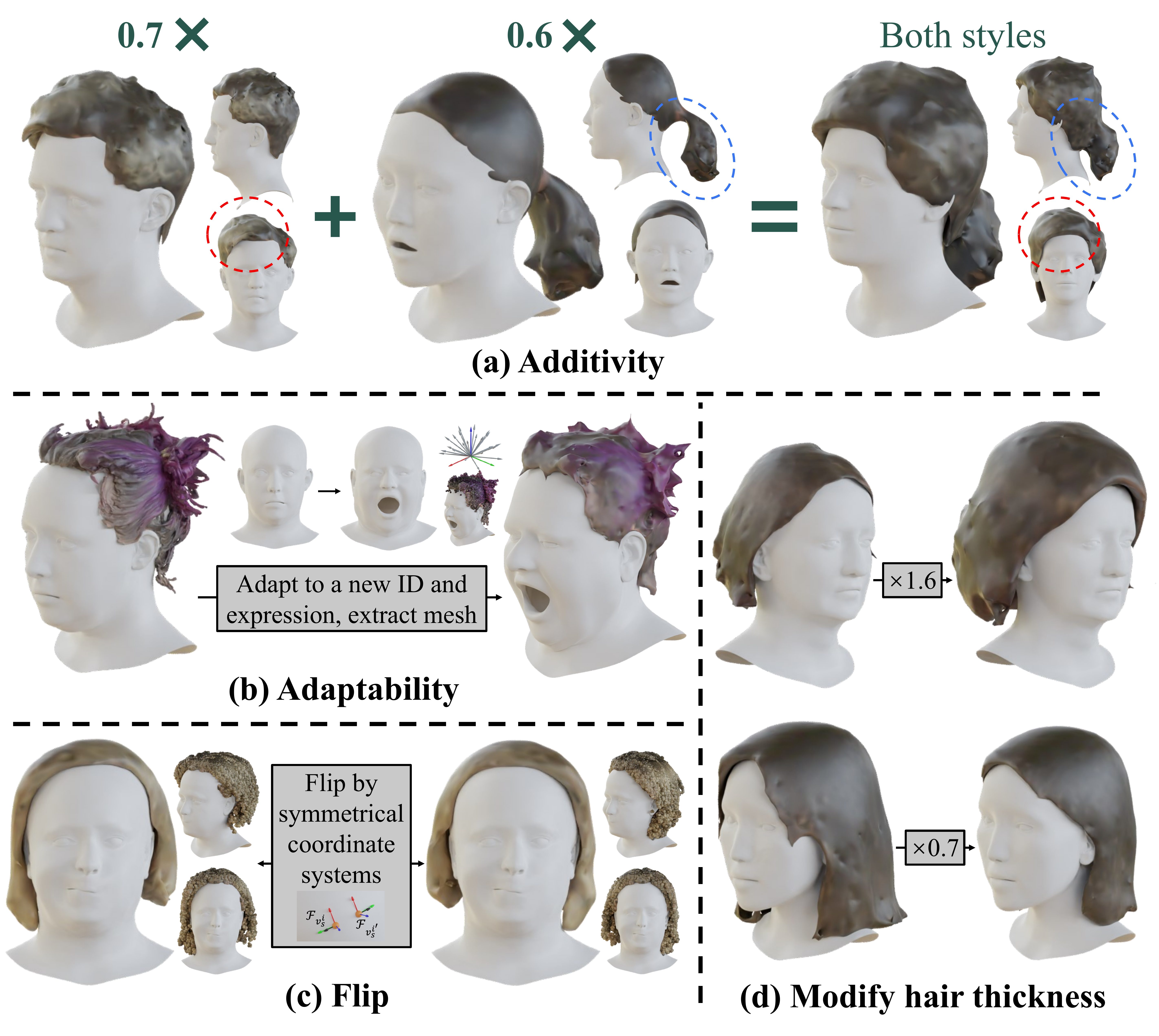}
\end{center}
\vspace{-0.5cm}
\caption{The properties of semantic-consistent ray modeling.
}
\label{properties}
 \vspace{-0.2cm}
\end{figure}

\myparagraph{Hair Vertex Statistics Using Shortest and Longest Ray Distances.} Based on the semantic-consistent ray sets on the scalp $\{\mathcal{R}_{{ray}}^i\}_{i=1}^{N_s}$, we record the distances from the origin $v_s^{i}$ of each ray $\mathbf{r}_n \in \mathcal{R}_{{ray}}^{i}$ to the vertices of the hair mesh ${{\cal M}}_{hair}$ that $\mathbf{r}_n$ intersects. We only record the nearest and farthest distances to ensure that the distances also have statistical properties. Since we have sufficient scalp vertices and each $\mathcal{R}_{{ray}}^i$ contains sufficient $N_r$ ray directions, we are able to achieve coverage and statistical analysis on complex meshes, as shown in Fig.~\ref{modeling}(c), resulting in the ray distance set $\mathcal{D}_{{ray}} = \{\mathcal{D}_{{\max}}, \mathcal{D}_{{\min}}\} \in \mathbb{R}^{N_s \cdot  N_r \cdot  2}$. The values $\mathcal{D}_{{\max}}(i,n)$ and $\mathcal{D}_{{\min}}(i,n)$ at the position $(i,n)$ of $\mathcal{D}_{{\max}} \in \mathbb{R}^{N_s \cdot  N_r}$ and $\mathcal{D}_{{\min}} \in \mathbb{R}^{N_s \cdot  N_r}$ are determined by:
\begin{equation}{
\left\{
\begin{aligned}
\mathcal{D}_{{\max}}(i,n) = \max_{v \in {{\cal M}}_{hair}} \left\| (v - v_s^{i}) \cdot \mathbf{r}_n \right\|_2\\
\mathcal{D}_{{\min}}(i,n) = \min_{v \in {{\cal M}}_{hair}} \left\| (v - v_s^{i}) \cdot \mathbf{r}_n \right\|_2
\end{aligned}\text{   .}
\right.
}
\label{distance-equ}
\end{equation}

 To simplify the notation, we use $d_{\max}$ and $d_{\min}$ in Fig.~\ref{modeling}(c) to represent $\mathcal{D}_{{\max}}(i,n)$ and $\mathcal{D}_{{\min}}(i,n)$, respectively. There are three possible cases regarding the intersection of a ray $\mathbf{r}_n$ with the hair mesh $\mathcal{M}_{{hair}}$, as illustrated in Fig.~\ref{modeling}(c): \textit{{Case 1:}} If the ray $\mathbf{r}_n$ emitted from $v_s^i$ does not intersect any vertex of the hair mesh, set $d_{\max} = d_{\min} = 0$. \textit{{Case 2:}} If the ray $\mathbf{r}_n$ intersects the hair mesh $\mathcal{M}_{{hair}}$ at exactly one point, set $d_{\max} = d_{\min}$. \textit{{Case 3:}} If the ray $\mathbf{r}_n$ has multiple intersection points with the hair mesh $\mathcal{M}_{{hair}}$, the values are determined according to Eqn.~\ref{distance-equ}, \ie, record the nearest and farthest distances. Additionally, while obtaining $\mathcal{D}_{{ray}}$, we simultaneously record the corresponding albedo associated with each ray's intersections $
\mathcal{A}_{{hair}} \in \mathbb{R}^{3 \cdot  (N_s \cdot  N_r \cdot  2)}$. In \textit{Case 1}, set the corresponding value in $\mathcal{A}_{{hair}}$ to the mean texture of the skin.

Note that our hair vertex statistics are performed based on the ray distances rather than the coordinates of the hair vertices, ensuring the invariance to head pose variations. As shown in Fig.~\ref{modeling}(d), the vertices on the hair surface $V_{{hair}} \in \mathbb{R}^{3 \cdot  (N_s \cdot  N_r \cdot  2)}$ can be represented by:
\begin{equation}
\begin{aligned}
\begin{array}{l}
V_{{hair}} = \left\{ v_s^i + \mathbf{r}_n \cdot \mathcal{D}_k(i,n)  \right\}_{(i,n,k)}
\end{array}
\end{aligned},
\label{v_hair-equ}
\end{equation}
where $v_s^i$ is the scalp vertex, $\mathbf{r}_n$ is the ray direction, and $\mathcal{D}_k(i,n)$ represents the corresponding ray distance in Eqn.~\ref{distance-equ}, with $k \in \{\min, \max\}$. $i $=$ 1, \ldots, N_s  $. $n $=$ 1, \ldots, N_r $. With the assistance of our semantic-consistent ray modeling, the ordered element sets $\mathcal{D}_{{ray}}$, $V_{{hair}}$, and $\mathcal{A}_{{hair}}$ could be obtained. If necessary, users can further filter out the scalp vertices in $V_{{hair}}$ to obtain the final result.

\begin{algorithm}[t]
\footnotesize
\footnotesize\caption{\footnotesize{Get Mesh Triangle Faces from Vertices}}
\label{algo:get_faces_kaolin}
\SetAlgoLined
\SetKwInput{KwData}{Input}
\SetKwInput{KwResult}{Output}
\KwData{Batch vertices: $\mathcal{V} \in \mathbb{R}^{B \cdot 3 \cdot N}$, Voxel size: $\Delta$}
\KwResult{The mesh face list for each $\mathcal{V}_b \in \mathcal{V}$: $\mathcal{T}$}

\For{$b = 1$ \KwTo $B$}{
    $\mathcal{V}_b^{'}, scale_b, trans_b \gets \text{Normalize}(\mathcal{V}_b)$\;
    
    $\mathcal{P}_b^{'} \gets \text{PointToVoxelGrid}(\mathcal{V}_b^{'}, \Delta)$\;
    
    \tcp{Get faces $\mathcal{P}_b^{f}$ from $\mathcal{P}_b^{'}$ }
    $(\mathcal{P}_b^{'}, \mathcal{P}_b^{f}) \gets \text{VoxelToMesh}(\mathcal{P}_b^{'})$\;

    $\mathcal{P}_b \gets \text{Re-normalize}(\mathcal{P}_b^{'}, scale_b, trans_b$)\;
    
    \tcp{KNN search to match $\mathcal{V}_b$ with $\mathcal{P}_b$}
    $\mathcal{K}_b \gets \text{KnnSearch}(\mathcal{P}_b, \mathcal{V}_b)$\;

     \tcp{Filter out far neighbors and transform valid face indices for $\mathcal{V}_b$}
    $\mathcal{T}_b \gets \text{Filter}(\mathcal{P}_b^{f}, \mathcal{K}_b)$\;
}

\vspace{-1mm}
\end{algorithm}

\subsection{Properties of the Modeling} 
Semantic-consistent Ray Modeling for Hair (SRM-Hair) exhibits several notable properties. As shown in Fig.~\ref{properties}(a), its additivity allows for the combination of a short hairstyle with dense hair on the right side and a ponytail, producing a final 3D hair mesh that integrates features from both styles, controlled by specific coefficients.\textbf{ Additivity serves as the foundation of morphable hair}. Fig.~\ref{properties}(b) shows SRM-Hair's ability to transfer complex known hair data to another head with a new identity and expression. Fig.~\ref{properties}(c) and (d) showcase that SRM-Hair can support flipping hairstyles and modifying hair thickness. To the best of our knowledge, SRM-Hair is the first hair modeling method to simultaneously exhibit these capabilities. The mesh extraction process in Fig.~\ref{properties} is described in the next section.

\section{Reconstruction}
\label{4Reconstruction}
\begin{figure*}[t]
\begin{center}
\includegraphics[width=1\linewidth]{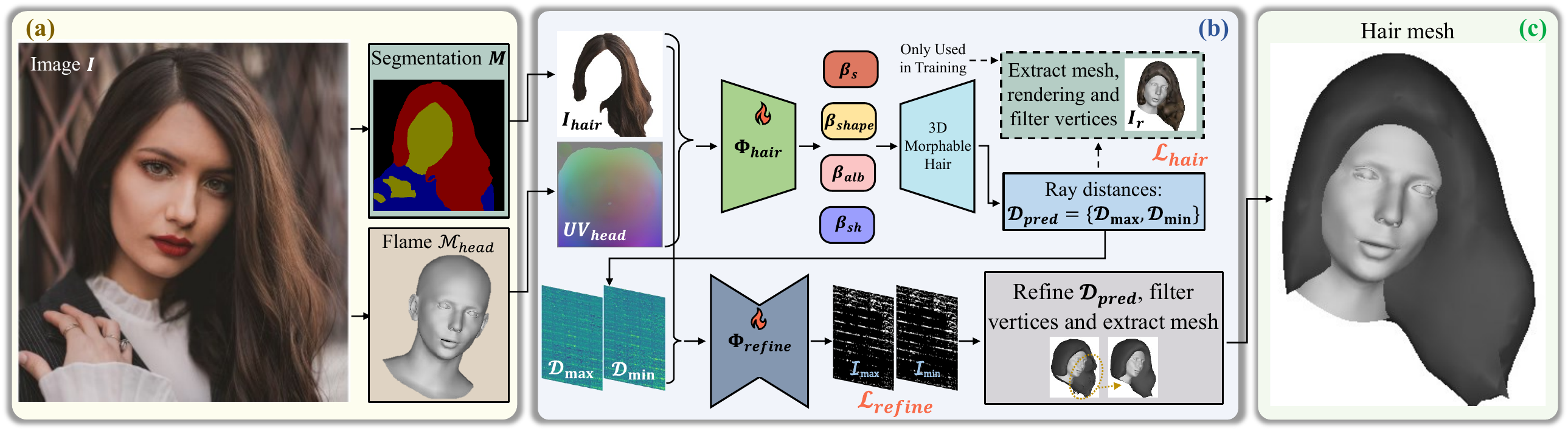 }
\end{center}
\vspace{-0.5cm}
\caption{Overview of SRM-Hair for reconstructing a 3D hair mesh from a single image.}
\label{reconstruction}
\vspace{-0.5cm}
\end{figure*}

\subsection{3D Morphable Hair and Preliminaries}

\myparagraph{3D Morphable Hair.} To ensure the robustness of the 3D morphable hair, we flip each sample in the hair dataset, resulting in over 500 samples. Based on additivity and the hair dataset, we perform Semantic-consistent Ray Modeling on the $ N_{{sample}} $ annotated hair meshes to obtain $\mathcal{D}_{{ray}}^{{all}} \in \mathbb{R}^{N_{{sample}} \cdot  N}$ and $\mathcal{A}_{{hair}}^{{all}} \in \mathbb{R}^{N_{{sample}} \cdot  3N}$, where $N = N_s \cdot  N_r \cdot  2$, $ N_{{sample}} = 480 $ ({\ie} 240 hair samples and their corresponding flipped data are used), $N_s = 900$, and $N_r = 25$. Then, we obtain a 3D morphable hair that can represent a new hairstyle through coefficients based on \cite{wood2021fake, blanz1999morphable} and Principal Component Analysis (PCA) \cite{Sirovich:87, jolliffe2016principal}:
\begin{equation}
\begin{aligned}
\begin{array}{l}
{\mathcal{D}_{pred}} =  ({\overline{\mathcal{D}}}  + {{\beta} _{shape}}{{\mathcal{D}}_{shape}}) \cdot {\beta} _{s} \\
{\mathcal{A}_{pred}} = {\overline{\mathcal{A}}}  + {{\beta} _{alb}}{\mathcal{A}_{alb}}
\end{array}
\end{aligned},
\label{pca_ray-equ}
\end{equation}
where $\mathcal{D}_{pred} \in \mathbb{R}^{N}$ is the predicted ray distances and ${\mathcal{A}_{pred}\in {{\mathbb{R}}^{3 \cdot {N}}}}$ is the predicted albedo. $N = N_s \cdot  N_r \cdot  2$. ${\overline{\mathcal{D}}} \in \mathbb{R}^{N}$ and ${\overline{\mathcal{A}}} \in \mathbb{R}^{3 \cdot N}$ are the mean ray distances the mean albedo calculated by $\mathcal{D}_{{ray}}^{{all}}$ and ${\mathcal{A}}_{{hair}}^{{all}} $, respectively. 
${{\mathcal{D}}_{shape}}$ and ${\mathcal{A}_{alb}}$ are the ray distance bases and the albedo bases, respectively, obtained by applying PCA \cite{Sirovich:87, jolliffe2016principal,wood2021fake, blanz1999morphable} to $\mathcal{D}_{{ray}}^{{all}}$ and ${\mathcal{A}}_{{hair}}^{{all}} $. ${{{\beta} _{shape}} \in {{\mathbb{R}}^{{480}}}}$ and ${{{\beta} _{alb}} \in {{\mathbb{R}}^{{480}}}}$ are the ray distance coefficients and the albedo coefficients, respectively. ${{{\beta} _{s}} \in {{\mathbb{R}}^+}}$ controls the scale of ray distances, modifying the hair thickness as shown in Fig.~\ref{properties}(d). Consistent with Eqn.~\ref{distance-equ}, $\mathcal{D}_{pred}$ can also be divided into $\mathcal{D}_{\max}$ and $\mathcal{D}_{\min}$, {\ie}, $\mathcal{D}_{pred} = \{{\mathcal{D}_{\max}, \mathcal{D}_{\min}}\}$.

Based on the obtained ray distances $\mathcal{D}_{pred}$, given a FLAME mesh ${{\cal M}}_{{head}}$ with arbitrary pose, expression, and identity, we first obtain the scalp $V_s$ from ${{\cal M}}_{{head}}$ and compute the corresponding semantic-consistent ray sets $\mathcal{R}_{{ray}}^{{scalp}}$. Then, we use Eqn.~\ref{v_hair-equ} to compute the new hair vertices $V_{pred} \in \mathbb{R}^{3 \cdot N}$. Since the hair pose is determined by the provided head ${{\cal M}}_{{head}}$, the geometry of our 3D morphable hair does not include rotation and translation, and only considers the length of the rays as shown in Eqn.~\ref{pca_ray-equ}.

\myparagraph{Mesh Extraction.} SRM-Hair leverages existing 3D processing techniques \cite{KaolinLibrary} to convert the vertices into a corresponding mesh, as shown in Algo.~\ref{algo:get_faces_kaolin}, which ensures that the generated hair vertices $V_{pred}$ can be converted into a mesh with triangle-faces $tri$, supports batch input, and integrated with advanced differentiable rendering techniques \cite{ravi2020pytorch3d, KaolinLibrary, chen2019learning, Laine2020diffrast}. When differentiability is not required, the algorithm can also be combined with \cite{vollmer1999improved, trimesh, Zhou2018open3d} to smooth $V_{pred}$ for improved visual quality.

\myparagraph{Camera, Illumination and Rendering.} Based on \cite{DECA:Siggraph2021,SMIRK}, we use an orthographic
camera for the re-projection of ${{V_{pred}}}$ into the 2D image plane, resulting ${{V_{pred}^{2d}} \in {{\mathbb{R}}^{2 \cdot {N}}}}$. Following \cite{DECA:Siggraph2021, deng2019accurate, wang2024s2td}, we utilize Spherical Harmonics (SH) \cite{ramamoorthi2001efficient} to estimate shading: 
\begin{equation}
\small
\begin{aligned}
\begin{array}{l}
S({\beta}_{sh} ,{\mathcal{A}_{pred}},{\mathbf{n}_{pred}}) = {\mathcal{A}_{pred}} \odot \sum\limits_{k = 1}^9 {{\beta}_{sh}^k} {{{\Psi }}_k}({\mathbf{n}_{pred}})
\end{array}
\end{aligned},
\end{equation}
where $S({\beta}_{sh} ,{\mathcal{A}_{pred}},{\mathbf{n}_{pred}})$ is the predicted texture. $\odot$ represents the Hadamard product, ${\mathbf{n}_{pred}}$ denotes the normal of ${{V_{pred}}}$, and ${\Psi}: \mathbb{R}^{3} \rightarrow \mathbb{R}$ is the SH basis function. The coefficients ${{\beta}_{sh}^k} \in \mathbb{R}^3$ corresponds to the SH parameter for each $k \in \{1, \cdots, 9\}$. ${\mathcal{A}_{pred}}$ is the albedo in Eqn.~\ref{pca_ray-equ}. The differentiable rendering of the predicted hair $I_r$ is:
\begin{equation}
\begin{aligned}
\begin{array}{l}
I_r = \text{Render}(V_{pred},tri,S({\beta}_{sh} ,{\mathcal{A}_{pred}},{\mathbf{n}_{pred}}))
\end{array}
\end{aligned},
\label{render-equ}
\end{equation}
where $tri$ represents the triangle faces corresponding to the hair vertices $V_{pred}$ by using Algo.~\ref{algo:get_faces_kaolin}. We use the \textit{range mode} of Nvdiffrast \cite{Laine2020diffrast} to address the issue of varying triangle-faces in a batch.

\subsection{Reconstruction Framework}
\myparagraph{Preprocessing.} SRM-Hair employs RetinaFace \cite{Deng2020CVPR} to crop the face, resulting in an input RGB image ${{I} \in \mathbb{R}^{H \cdot W \cdot 3}}$. As shown in Fig.~\ref{reconstruction}(a), SRM-Hair firstly predicts the segmentation mask $M$ and the FLAME mesh ${{\cal M}}_{{head}}$ from $I$. The segmentation mask is represented as $M = \{ M_p \mid p \in P \}$, where $P = \{\text{hair}, \text{skin}, \text{clothes}, \text{background}\}$ and each $M_p \in \{0,1\}^{H \cdot W}$. A pixel at position $(x, y)$ satisfies $M_p^{(x,y)} = 1$ if it belongs to the part $p$. Following \cite{wang20243d}, $M$ is further converted into point sets during training, formulated as $C = \{ C_p \mid p \in P \}$, where $C_p = \{ (x, y) \mid M_p^{(x,y)} = 1 \}$. $C_{{hair}}$ further includes 2D scalp vertices $V_s^{2d}$ to match with $V_{pred}$ and guide its deformation during training. $M$ is from MediaPipe \cite{lugaresi2019mediapipe} and ${{\cal M}}_{{head}}$ is obtained by DECA \cite{DECA:Siggraph2021} or SMIRK \cite{SMIRK}. SRM-Hair transfers head geometry ${{\cal M}}_{{head}}$ to a UV position map ${UV}_{head}$ based on \cite{feng2018joint,DECA:Siggraph2021} and gets hair region image  ${I_{hair}}$ as the input of the next step.

\myparagraph{Predict Morphable Hair Coefficients.} As shown in the upper part of Fig.~\ref{reconstruction}(b), SRM-Hair uses ResNet-50 \cite{he2016deep} as the backbone ${{\bf{\Phi }}_{{{hair}}}}$ to predict coefficients ${{\beta} _{s}}$, ${{\beta}_{shape}}$, ${{\beta} _{alb}}$, and ${{\beta} _{sh}}$ based on ${I_{hair}}$ and ${UV}_{head}$. The 3D morphable hair processes coefficients to generate ray distances $\mathcal{D}_{pred}$ and the albedo $\mathcal{A}_{pred}$, as described in Eqn.~\ref{pca_ray-equ}. The predicted hair vertices $ V_{{pred}} $ are determined by the head mesh ${{\cal M}}_{{head}}$ and the predicted ray distances $\mathcal{D}_{{pred}}$. During training, SRM-Hair obtains the predicted hair image $ I_r $ using Algo.~\ref{algo:get_faces_kaolin} and Eqn.~\ref{render-equ}. $ V_{{pred}} $ and $ I_r$ are filtered based on the mask $ M $ to align with the points $ C_{{hair}} $ and \( I_{\text{hair}} \). The simple vertex filtering process is detailed in the supplementary material. The predicted ray distances $\mathcal{D}_{\max}$ and $\mathcal{D}_{\min}$ are reshaped into $2 \cdot (150 \cdot 150)$ distance maps which are used as inputs for the next step.

\myparagraph{Refinement.} As shown in the lower part of Fig.~\ref{reconstruction}(b), SRM-Hair uses a pix2pix network \cite{isola2017image} ${{\bf{\Phi }}_{{refine}}}$ to predict the index exclusion maps $\mathcal{I} = \{{{{\mathcal{I}}}_{\max}},{{{\mathcal{I}}}_{\min}}\}$ based on $\mathcal{D}_{\max}$ and $\mathcal{D}_{\min}$. During inference, $\mathcal{I} \in \{0,1\}^{2 \cdot (150 \cdot 150)}$ is obtained by applying a sigmoid function and binarizing with a 0.5 threshold, where 1 indicates positions in $\mathcal{D}_{{pred}}$ corresponding to outlier distances to be removed, thereby refining $\mathcal{D}_{{pred}}$. Please refer to the supplementary material and Sec.~\ref{AblationStudy} for more analysis of ${\bf{\Phi }}_{{{refine}}}$. The final hair mesh reconstruction is shown in Fig.~\ref{reconstruction}(c).

\subsection{Training Strategies}
\label{subsec:TrainingStrategies}

To train ${{\bf{\Phi }}_{{{hair}}}}$ and ${{\bf{\Phi }}_{{{refine}}}}$, we utilize a combination of 3D scans and in-the-wild 2D images. For 2D in-the-wild images, SRM-Hair uses the Chamfer distance ${\mathcal{L}_{cd}}$ and the rendering-based photometric loss ${\mathcal{L}_{pho}}$. The 2D images are randomly rotated \cite{bradski2000opencv} and undergo random changes in saturation, hue, contrast, {\etc}, during training. ${\mathcal{L}_{2d}}$ for 2D images is determined by:
\begin{equation}
\small
\begin{aligned}
 {\mathcal{L}_{2d}} &= \lambda_{{cd}}^{2d}{\mathcal{L}_{cd}}({{V}_{pred}^{2d}},{C_{hair}}) + \lambda_{{pho}}{\mathcal{L}_{pho}}({I_{hair}},{I_r},{M_{hair}})  \\ 
  &= \lambda_{{cd}}^{2d} \frac{1}{|{V}_{{pred}}^{2d}|} \sum_{v \in {V}_{{pred}}^{2d}} \min_{c \in C_{{hair}}} \| v - c \|_2^2  \\
  & \quad  +\lambda_{{cd}}^{2d}  \frac{1}{|C_{{hair}}|} \sum_{c \in C_{{hair}}} \min_{v \in {V}_{{pred}}^{2d}} \| c - v \|_2^2 \\
& \quad +\lambda_{{pho}}  \big\| {M_{hair}} \cdot(I_{hair}-I_{r})  \big\|_2^2
\end{aligned},
\end{equation}
where $V_{{pred}}^{2d}$ and $I_r$ are the predicted 2D hair vertices and the rendered image. $C_{{hair}}$ and $I_{{hair}}$ are their ground truths. The mask $M_{{hair}}$ make ${\mathcal{L}_{pho}}$ focus on the hair region. $\lambda_{{cd}}^{2d}$ and $\lambda_{{pho}}$ are the corresponding weights. 

For 3D data, SRM-Hair uses renderers \cite{ravi2020pytorch3d,Laine2020diffrast,vhap} to generate 2D images with varying poses and lighting as inputs and employs semantic-consistent ray modeling to obtain the ground truths $\mathcal{D}_{{gt}}$, $V_{{gt}}$ and ${\mathcal{A}_{{gt}}}$ for element-wise comparison with the predictions $V_{{pred}}$ and ${\mathcal{A}_{{pred}}}$ . SRM-Hair also uses the Chamfer distance for 3D data. The overall 3D loss ${\mathcal{L}_{3d}}$ is as follows:
\begin{equation}
\small
\begin{aligned}
 {\mathcal{L}_{3d}}  
  &= \lambda_{{cd}}^{3d} \frac{1}{|{V}_{{pred}}|} \sum_{v \in {V}_{{pred}}} \min_{v' \in V_{{gt}}} \| v - v' \|_2^2  \\
   &\quad +\lambda_{{cd}}^{3d}  \frac{1}{|V_{{gt}}|} \sum_{v' \in V_{{gt}}} \min_{v \in {V}_{{pred}}} \|  v' - v \|_2^2 \\
  &\quad+\lambda_{{v}}  \big\|  {V}_{{pred}} - {V}_{{gt}} \big\|_2^2+\lambda_{{alb}}  \big\|  {\mathcal{A}_{{pred}}} - {\mathcal{A}_{{gt}}} \big\|_2^2
\end{aligned},
\end{equation}
where $\lambda_{{cd}}^{3d}$, $\lambda_{{v}}$, and $\lambda_{{alb}}$ are the corresponding weights. The total loss ${\mathcal{L}_{hair}}$ to train ${{\bf{\Phi }}_{{{hair}}}}$ is:
\begin{equation}
\begin{aligned}
 {\mathcal{L}_{hair}}  
  = \lambda_{{2d}} {\mathcal{L}_{2d}}  
 +\lambda_{{3d}} {\mathcal{L}_{3d}}   + \lambda_{{reg}}{\mathcal{L}_{reg}}
\end{aligned},
\end{equation}
where ${\mathcal{L}_{reg}}$ is the regularization loss following \cite{wang20243d,deng2019accurate}. $\lambda_{{2d}}$, $\lambda_{{3d}}$, and $\lambda_{{reg}}$ are the corresponding weights. During training, 2D and 3D data are mixed, with a function for selecting ${\mathcal{L}_{2d}}$ or ${\mathcal{L}_{3d}}$ in a batch.

To train ${{\bf{\Phi }}_{{refine}}}$, we construct the predicted index exclusion maps $\mathcal{I}$ and their corresponding ground truth $\mathcal{I}_{{gt}}$, while recording the input $\mathcal{D}_{{pred}}$ for ${{\bf{\Phi }}_{{refine}}}$. For 2D data, we manually annotate over 150 samples. Specifically, for each prediction from ${{\bf{\Phi }}_{{hair}}}$, we mark the outlier vertices that need removal to build $\mathcal{I}_{{gt}}$. For 3D data, since we have $\mathcal{D}_{{gt}}$, we perturb data at arbitrary positions of $\mathcal{D}_{{gt}}$ during training and record these positions to construct $\mathcal{I}_{{gt}}$.  Based on this, we train ${{\bf{\Phi }}_{{refine}}}$ using binary cross-entropy (BCE) loss \cite{ruby2020binary}:
\begin{equation}
\begin{aligned}
\mathcal{L}_{{refine}} &= - \mathcal{I}_{{gt}} \cdot \log\big(\sigma(\mathcal{I})\big) \\
& \quad - (1 - \mathcal{I}_{{gt}}) \cdot \log\big(1 - \sigma(\mathcal{I})\big)
\end{aligned},
\end{equation}
where $\sigma(\cdot)$ is the sigmoid function.

\begin{figure*}[t]
\begin{center}
\includegraphics[width=1\linewidth]{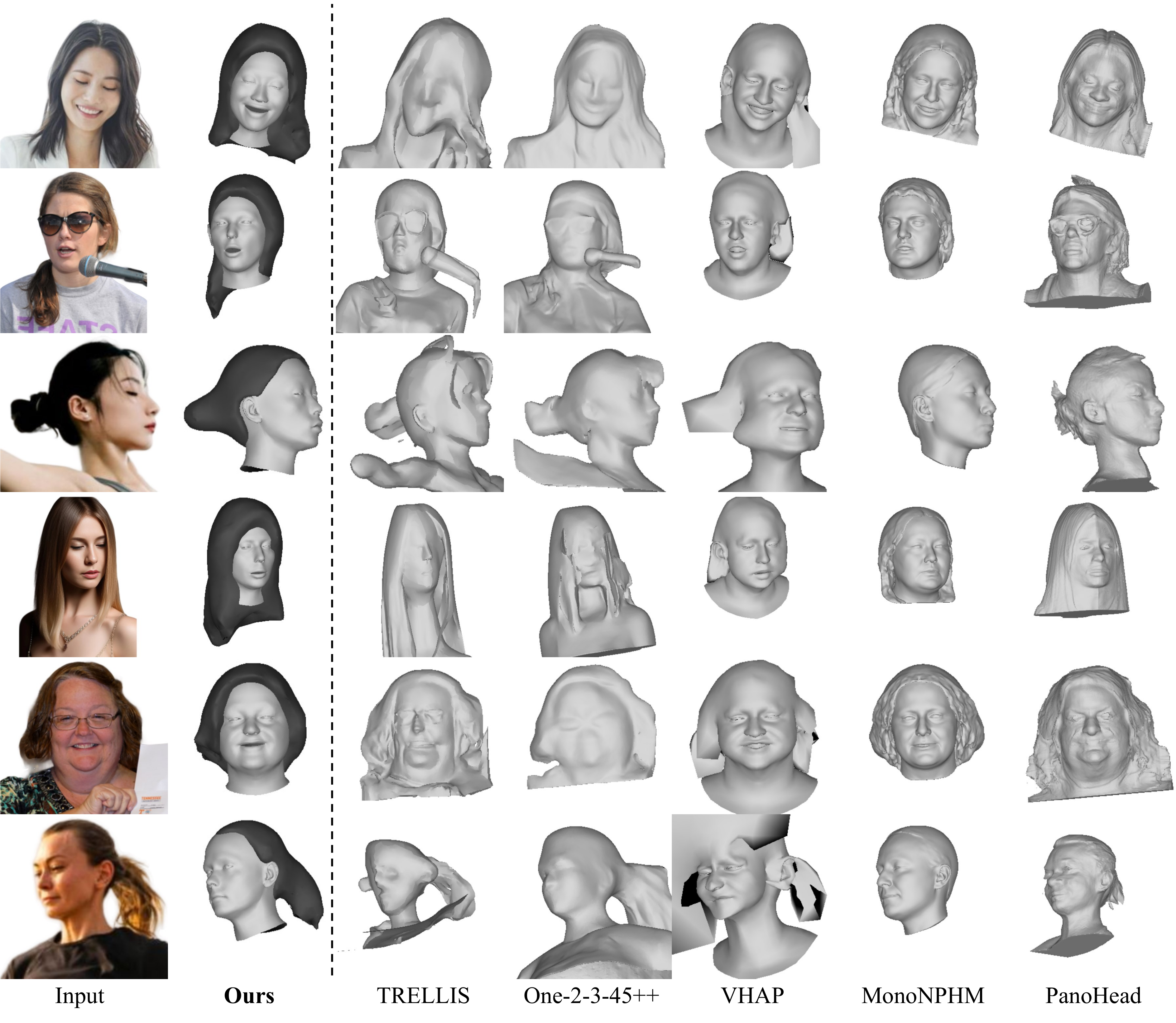 }
\end{center}
\vspace{-0.8cm}
\caption{Qualitative comparison with the other methods. Our method (SRM-Hair) achieves the best results, accurately reflecting the full head region. Furthermore, our reconstructed hair (shown in dark gray) and face are separated, enhancing the potential for application.
}
\label{compare}
\vspace{-0.4cm}
\end{figure*}

\section{Experiments}

\subsection{Experimental Settings}
\myparagraph{Training Data.} To train ${{\bf{\Phi }}_{{{hair}}}}$ and ${{\bf{\Phi }}_{{{refine}}}}$ in SRM-Hair, 3D datasets FaceVerse \cite{wang2022faceverse} and NPHM \cite{giebenhain2023nphm} (240 identities, 480 samples with flip) are used, along with 2D in-the-wild images from FFHQ \cite{karras2019style} and online collections (total over 10K images). With data augmentation mentiond in Sec.~\ref{subsec:TrainingStrategies}, over 500K data pairs are used to train ${{\bf{\Phi }}_{{hair}}}$. SRM-Hair trains ${{\bf{\Phi }}_{{hair}}}$ first, followed by ${{\bf{\Phi }}_{{refine}}}$. 

\myparagraph{Implementation Details.} We implement SRM-Hair based on PyTorch \cite{paszke2019pytorch}, Nvdiffrast \cite{Laine2020diffrast}, PyTorch3D \cite{ravi2020pytorch3d} and Kaolin \cite{KaolinLibrary}. The input hair image ${I_{hair}}$ are resized into $224\cdot 224$. We use Adam \cite{kingma2014adam} as the optimizer with an initial learning rate of $1e$ - $4$.

\myparagraph{SOTA Methods for Comparison.} We evaluated our methods with state-of-the-art mesh reconstruction approaches, including One-2-3-45++ \cite{liu2023one2345++}, MonoNPHM \cite{giebenhain2024mononphm}, PanoHead \cite{an2023panohead}, VHAP \cite{vhap}, and TRELLIS \cite{TRELLIS}. One-2-3-45++ \cite{liu2023one2345++} and TRELLIS \cite{TRELLIS} are representative 3D generation methods for high-quality 3D asset creation. MonoNPHM \cite{giebenhain2024mononphm} and PanoHead \cite{an2023panohead} are widely recognized recent works in head reconstruction. VHAP \cite{vhap} is a photometric optimization pipeline based on differentiable mesh rasterization, applied to head alignment and widely used in recent 3DGS avatar tasks \cite{qian2024gaussianavatars}. Additionally, methods like UniHair \cite{unihair} focus more on hair rendering than mesh reconstruction. These types of methods tend to fail in wild scenarios and cannot generate reliable 3D meshes, which have different applications from ours. A detailed discussion and comparison are provided in the supplemental material.

\begin{table*}
\scriptsize
\caption{Quantitative comparison of 3D head reconstruction for the hair region and the full head region. The best is highlighted in bold.
}
\vspace{-18pt}
\begin{center}
\renewcommand{\arraystretch}{1.0}
\resizebox{1\textwidth}{!}
{

\begin{tabular}{c|cccccccc|cc}
\toprule[1pt]
\multirow{3}{*}{Method} & \multicolumn{8}{c|}{Hair Region}                                                                                                             & \multicolumn{2}{c}{Full Head}                                                                 \\ \cline{2-11} 
                        & \multicolumn{4}{c|}{NRMSE $\downarrow$}                                         & \multicolumn{4}{c|}{Recall $\uparrow$}                     & \multicolumn{1}{c|}{\multirow{2}{*}{NRMSE $\downarrow$}} & \multirow{2}{*}{Recall $\uparrow$} \\
                        & \textit{10K-GT} & \textit{50K-GT} & \textit{100K-GT} & \multicolumn{1}{c|}{\textit{Avg.}} &\textit{ 10K-GT} & \textit{50K-GT} & \textit{100K-GT} & \textit{Avg.} & \multicolumn{1}{c|}{}                                    &                                    \\ 
\midrule[0.75pt]

TRELLIS \cite{TRELLIS}                  & 0.04363 & 0.04384 & 0.04388  & \multicolumn{1}{c|}{0.04378}                         & 0.37365 & 0.37569 & 0.37637  & 0.37524                        & \multicolumn{1}{c|}{0.03084}                              & 0.59207                            \\

One-2-3-45++ \cite{liu2023one2345++}            & 0.03885 & 0.03916 & 0.03934  & \multicolumn{1}{c|}{0.03912}                         & 0.44049 & 0.43823 & 0.43617  & 0.43830                         & \multicolumn{1}{c|}{0.02858}                              & 0.61833                            \\

VHAP \cite{vhap}                      & 0.04375 & 0.04471 &  0.04500 & \multicolumn{1}{c|}{0.04449}                         & 0.47875 & 0.47149 & 0.46785  & 0.47270                         & \multicolumn{1}{c|}{0.03188}                              & 0.64997                          \\

MonoNPHM \cite{giebenhain2024mononphm}                & 0.07662 & 0.07869 & 0.07965  & \multicolumn{1}{c|}{0.07832}                         & 0.23179 & 0.22145 & 0.21477  & 0.22267                       & \multicolumn{1}{c|}{0.05356}                              & 0.50588                            \\

PanoHead \cite{an2023panohead}                 & 0.04147 & 0.04222 & 0.04244  & \multicolumn{1}{c|}{0.04204}                         & 0.46060 & 0.45536 & 0.45253 & 0.45616                        & \multicolumn{1}{c|}{0.03287}                              & 0.60689                            \\

\textbf{Ours }                   & \textbf{0.01636} & \textbf{0.01640} & \textbf{0.01636}  & \multicolumn{1}{c|}{\textbf{0.01637}}                         & \textbf{0.78870} & \textbf{0.78898} & \textbf{0.79002}  & \textbf{0.78923}                         & \multicolumn{1}{c|}{\textbf{0.01090}}                              & \textbf{0.86913  }                          \\
\bottomrule[1pt]
\end{tabular}

}
\end{center}
\vspace{-0.8cm}
\label{Quantitativecomparison}
\end{table*}

\myparagraph{Metrics.} We use five accurate hair scans from NPHM \cite{giebenhain2023nphm} paired with 3D faces as the 3D geometry ground truth (GT), where each sample is rendered into four images from different angles and lighting for testing (Total 20 samples). Note that these samples are not involved in SRM-Hair's 3D morphable hair or training process. We initially align the predictions and ground truth using face landmarks. Similar to \cite{chai2022realy}, we measure the reconstruction accuracy for both the hair region and the full head region using Normalized Root Mean Square Error (NRMSE). Note that methods such as \cite{liu2023one2345++, TRELLIS} cannot output an independent hair region, so simply using Chamfer distance for quantitative comparison would be unfair. The triangle number in the meshes from methods like \cite{liu2023one2345++, TRELLIS, giebenhain2024mononphm, an2023panohead, vhap} range from 10K to over 100K. To ensure a fair comparison, we evaluate the reconstruction accuracy using three versions of the ground truth mesh, remeshed \cite{garland1998simplifying} at different resolutions: 10K, 50K, and 100K triangles. This allows us to assess the robustness of each method across varying levels of mesh complexity. We also measure Recall \cite{tatarchenko2019single, giebenhain2024mononphm}, {\ie}, the percentage of ground truth points covered by at least one point in the reconstruction within a given threshold. We use Chamfer distance in the ablation study to quantify the impact of ${\bf{\Phi }}_{{{hair}}}$, ${\bf{\Phi }}_{{{refine}}}$, $\mathcal{D}_{\min}$, and $\mathcal{D}_{\max}$, as our method can independently output 3D hair mesh reconstructions.

\subsection{Quantitative Comparison}
\label{subsec:QuantitativeComparison}

The quantitative comparison is shown in Tab.~\ref{Quantitativecomparison}, indicating that our method achieved the best results on both hair region (NRMSE: $0.01637$, Recall: $0.78923$) and full Head region (NRMSE: $0.01090$, Recall: $0.86913$). SRM-Hair consistently outperform SOTA methods \cite{an2023panohead, liu2023one2345++, giebenhain2024mononphm, vhap, TRELLIS} in 3D geometry, demonstrating the effectiveness of our semantic-consistent ray modeling and the reconstruction method.

\begin{table}[t]
\caption{Ablation study for the impact of ${\bf{\Phi }}_{{{hair}}}$, ${\bf{\Phi }}_{{{refine}}}$, $\mathcal{D}_{\min}$, and $\mathcal{D}_{\max}$. The best is highlighted in bold.
}
\vspace{-0.6cm}
\begin{center}
{
\renewcommand{\arraystretch}{1.0}
\resizebox{0.39\textwidth}{!}
{
\begin{tabular}{cccc|c}
\toprule[1pt]
${\bf{\Phi }}_{{{hair}}}$ & ${\bf{\Phi }}_{{{refine}}}$ & $\mathcal{D}_{\min}$ & $\mathcal{D}_{\max}$ & Chamfer Distance $\downarrow$ \\ \midrule[0.75pt]
\textcolor{black}{\ding{51}} & \textcolor{gray!60}{\ding{55}}  &\textcolor{black}{\ding{51}}  & \textcolor{gray!60}{\ding{55}} &  0.04726 \\
\textcolor{black}{\ding{51}} & \textcolor{gray!60}{\ding{55}}  &\textcolor{gray!60}{\ding{55}}  & \textcolor{black}{\ding{51}} & 0.04219 \\
\textcolor{black}{\ding{51}} & \textcolor{gray!60}{\ding{55}}  &\textcolor{black}{\ding{51}}  & \textcolor{black}{\ding{51}} & 0.04061\\
\textcolor{black}{\ding{51}} & \textcolor{black}{\ding{51}}   &\textcolor{black}{\ding{51}}  & \textcolor{black}{\ding{51}} & \textbf{0.02446}\\ 
\bottomrule[1pt]
\end{tabular}
}
}
\end{center}
\label{ablation}
\vspace{-0.65cm}
\end{table}

\subsection{Qualitative Comparison} 

The qualitative comparison in Fig.~\ref{compare} shows that SRM-Hair produces the best results, accurately capturing the full head region. Moreover, our reconstruction separates the hair and face, increasing its applicability. Note that our hair vertex statistics are based on ray distances related to scalp rather than the coordinates of the hair vertices. This design of recording distances makes SRM-Hair robust to head pose variations and enables it to accurately capture a wide range of hairstyles, including short, long, afro, and ponytail styles.

\subsection{Ablation Study}
\label{AblationStudy}
Similar to the quantitative comparison setting in Sec.~\ref{subsec:QuantitativeComparison}, we use Chamfer distance to evaluate prediction accuracy at different stages and different ray-distance modeling in SRM-Hair. The ablation study investigates the impact of ${\bf{\Phi }}_{{{hair}}}$, ${\bf{\Phi }}_{{{refine}}}$, $\mathcal{D}_{\min}$, and $\mathcal{D}_{\max}$. The first two components belong to architecture design, while the latter two fall under semantic-consistent ray modeling for morphable hair. Note that the Chamfer distance measure is applicable here because, unlike \cite{an2023panohead, liu2023one2345++, giebenhain2024mononphm, vhap, TRELLIS}, SRM-Hair can independently output 3D hair geometry in a head reconstruction.

\myparagraph{Impact of ${\bf{\Phi }}_{{{hair}}}$ and ${\bf{\Phi }}_{{{refine}}}$ in Architecture Design.} As shown in Tab.~\ref{Quantitativecomparison} and Fig.~\ref{compare}, the semantic-consistent ray modeling and 3D morphable hair used in SRM-Hair are effective for reconstructing a 3D head with an independent hair mesh from a single image. However, due to the complexity of 3D hair surfaces, our semantic-consistent ray modeling does not strictly enforce point-consistent semantic meaning across shapes. This may lead to outlier points when generating new hairstyles in ${{\bf{\Phi }}_{{hair}}}$. Therefore, ${{\bf{\Phi }}_{{refine}}}$ is necessary to identify and remove outlier distances in $\mathcal{D}_{{pred}}$ as a refinement step. The ablation study in Tab.~\ref{ablation} further validates this necessity.

\myparagraph{Impact of $\mathcal{D}_{\min}$ and $\mathcal{D}_{\max}$ in Semantic-consistent Ray Modeling.} As shown in Tab.~\ref{ablation}, recording either the nearest distance $\mathcal{D}_{\min}$ or the farthest distance $\mathcal{D}_{\max}$ alone provides a rough coverage of the 3D hair surface vertices. However, recording both $\mathcal{D}_{\min}$ and $\mathcal{D}_{\max}$ enables a more comprehensive capture of the entire hair region. When combined with ${\bf{\Phi }}_{{{hair}}}$ and ${\bf{\Phi }}_{{{refine}}}$, SRM-Hair achieves the best results.

\section{Conclusions}

This paper proposes a novel method to reconstruct a 3D head with an independent hair mesh from a single image named SRM-Hair. The key contribution lies in semantic-consistent ray modeling for morphable hair, which implements point-consistent semantic meaning across hair shapes with several properties such as additivity and adaptability. The advanced modeling approach makes SRM-Hair enable single-image hair mesh reconstruction as simple as 3D face reconstruction with 3DMMs. We expect that this modeling approach could contribute to advancements in the reconstruction and animation of human-related objects, such as hair and clothing. Additionally, this paper contributes a high-fidelity hair mesh scan dataset. Extensive experiments demonstrate that our method outperforms existing state-of-the-art methods.

{
    \small
    \bibliographystyle{ieeenat_fullname}
    \bibliography{hair}
}

\newpage

\appendix

\maketitlesupplementary

\renewcommand*{\thefigure}{S\arabic{figure}}
\renewcommand*{\thetable}{S\arabic{table}}

\section{Simple Vertex Filtering Process} 
During training and inference, we filter the predicted hair vertices \( V_{{pred}} \) based on segmentation information \( M = \{ M_{{hair}}, M_{{skin}}, M_{{clothes}}, M_{{background}} \} \) and face geometry \( \mathcal{M}_{{head}} \). Specifically, we compute the average depth \( z_{{mean}} \) of \( \mathcal{M}_{{head}} \) as a reference depth, assuming \( \mathcal{M}_{{head}} \) is oriented along the positive \( z \)-axis. If any vertex \( v_h \in V_{{pred}} \) corresponds to a skin region ({\ie} \( v_h \in M_{{skin}} \)) and satisfies \( z(v_h) > z_{{mean}} \), it indicates that \( v_h \) is an outlier located in front of the head and attached to the skin region, and is therefore filtered out. 

\section{More Analysis about ${\bf{\Phi }}_{{refine}}$} 

\begin{figure}[t]
\begin{center}
\includegraphics[width=1\linewidth]{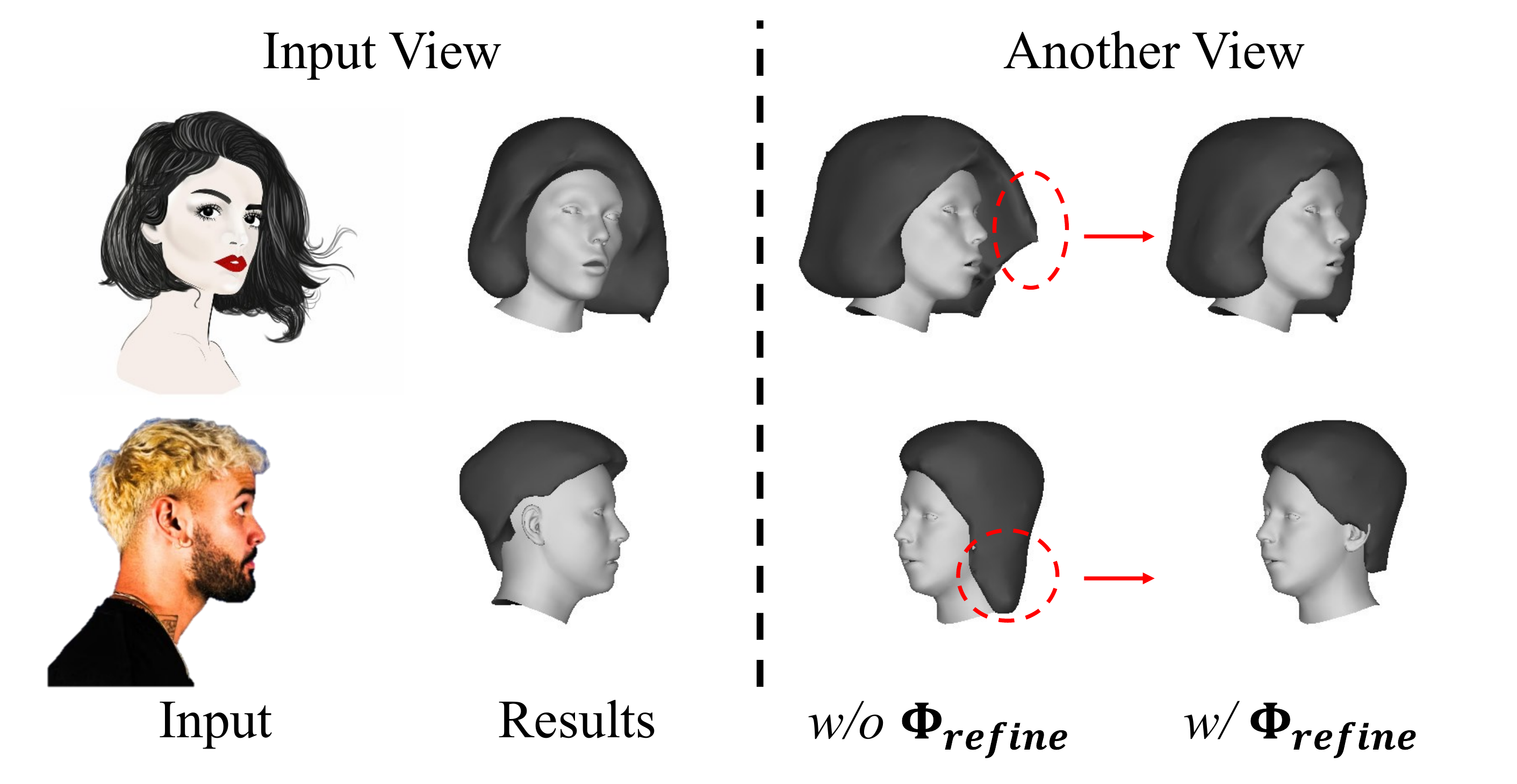}
\end{center}
\vspace{-0.5cm}
\caption{More analysis about ${\bf{\Phi }}_{{{refine}}}$. The introduction of ${\bf{\Phi }}_{{refine}}$ helps to further eliminate outliers generated by 3D morphable hair.
}
\label{sup_impactofrefine2}
 \vspace{-0.5cm}
\end{figure}

Tab. 1 and Fig. 5 in the main paper have shown the effectiveness of our semantic-consistent ray modeling and 3D morphable hair. However, due to the complexity of 3D hair surfaces, our semantic-consistent ray modeling does not strictly enforce point-consistent semantic meaning across different shapes. This may lead to outlier points when generating new hairstyles in ${\bf{\Phi }}_{{hair}}$. The introduction of ${\bf{\Phi }}_{{refine}}$ helps to further eliminate these outliers, as shown in Fig.~\ref{sup_impactofrefine2}.

\section{More Comparison with UniHair} 

\begin{figure}[t]
\begin{center}
\includegraphics[width=1\linewidth]{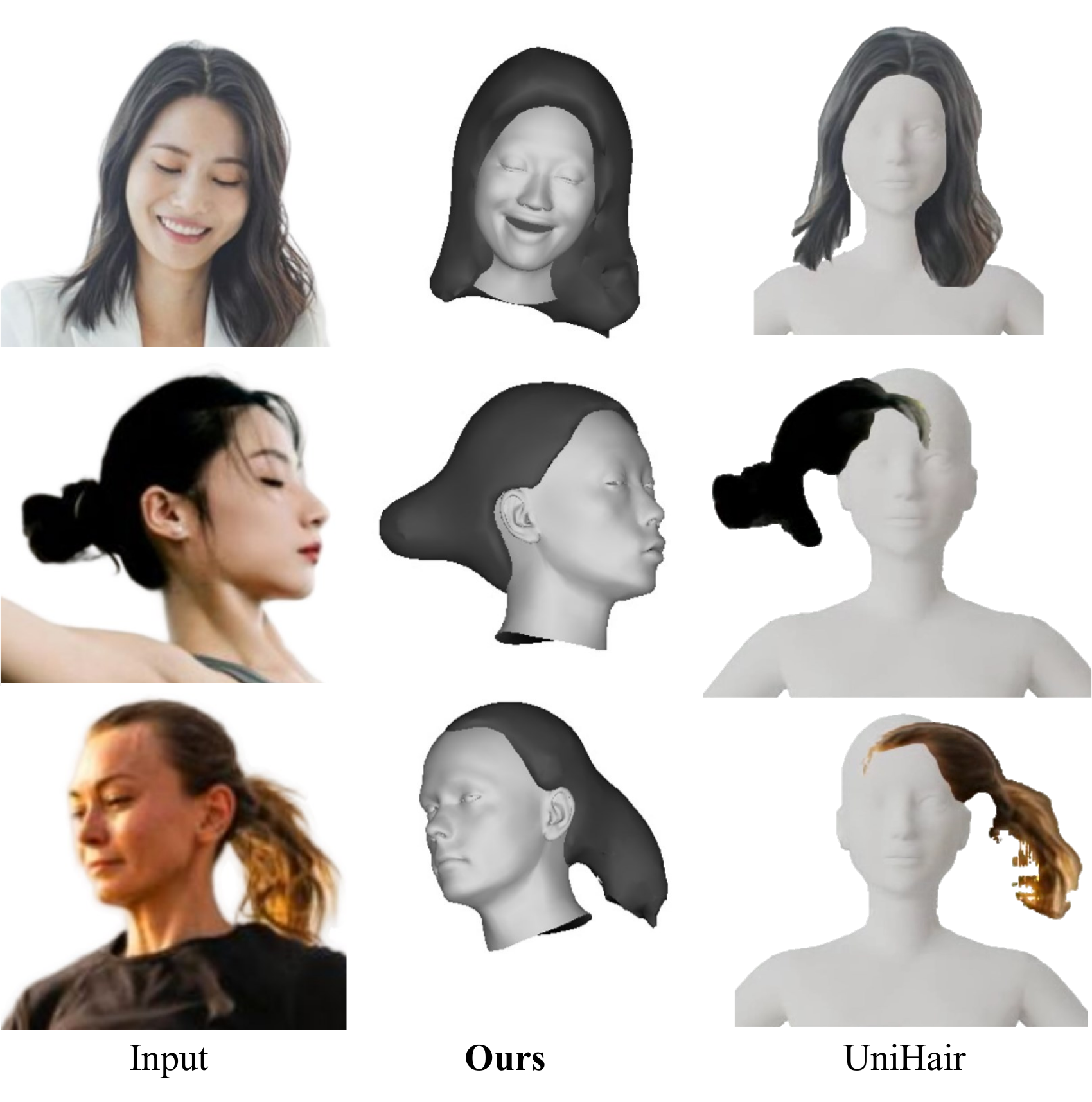}
\end{center}
\vspace{-0.6cm}
\caption{More comparison with UniHair \cite{unihair}.
}
\label{sup_unihair}
 \vspace{-0.5cm}
\end{figure}

Methods such as UniHair \protect\citesupp{unihairsupp} focus more on hair rendering rather than head geometry reconstruction, making them less capable of handling head pose variations in real-world scenarios. As shown in Fig.~\ref{sup_unihair}, we test several examples using the official UniHair \protect\citesupp{unihairsupp} code and found that UniHair \protect\citesupp{unihairsupp} struggles to produce accurate results under non-frontal views. Accurate estimation of hair region geometry enables 3D avatars to achieve more refined hair strand reconstruction \protect\citesupp{wu2024monohairsupp} or relighting effects \protect\citesupp{saito2024rgcasupp}. However, even with realistic rendering techniques, the resulting avatar may still appear distorted if the geometry is not precisely captured.

\section{More Results of SRM-Hair}

\begin{figure*}[t]
\begin{center}
\includegraphics[width=1\linewidth]{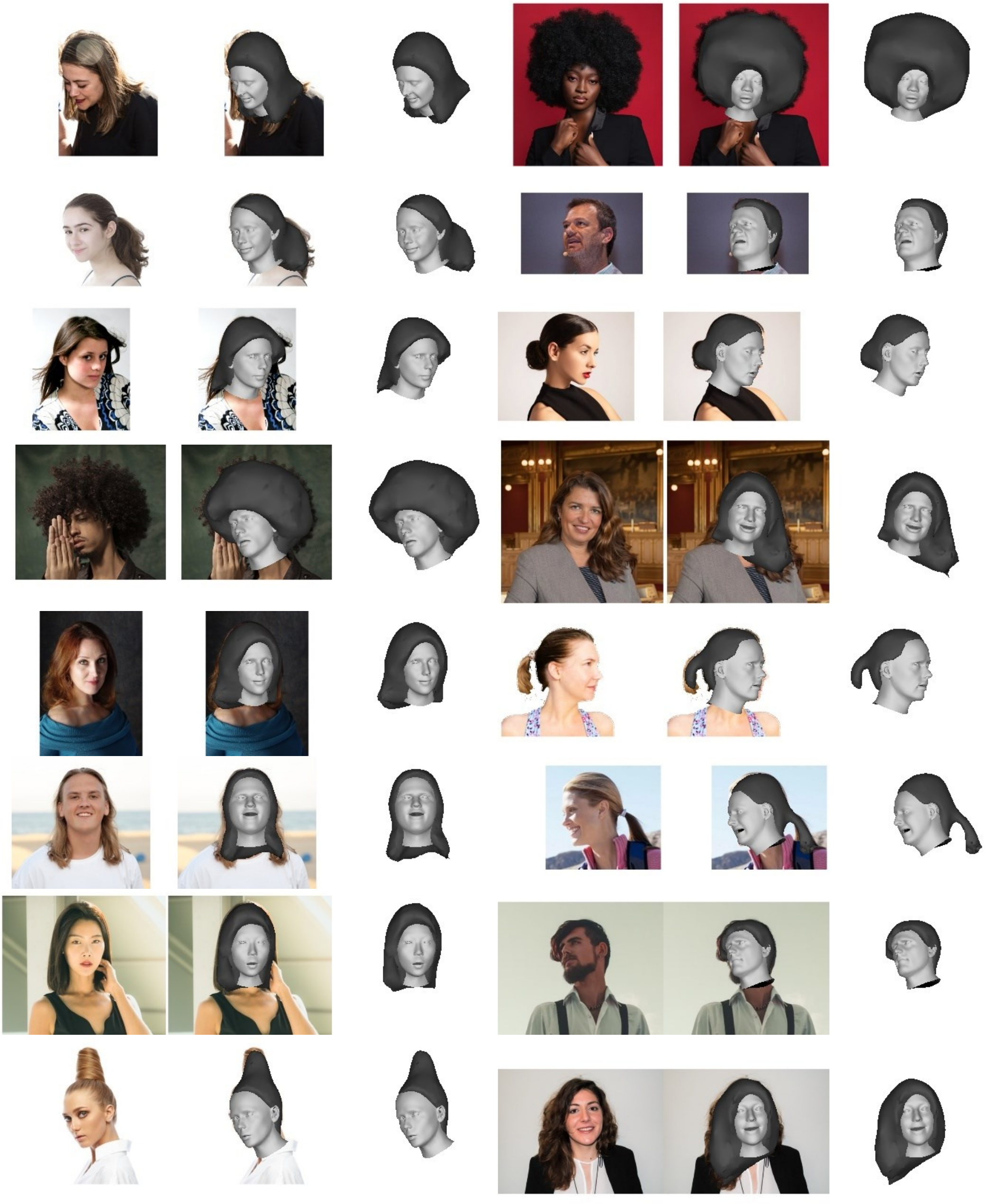 }
\end{center}
\vspace{-0.8cm}
\caption{Our method (SRM-Hair) accurately reflects the full
head region. The reconstructed hair and face of SRM-Hair are separated, enhancing the potential for application.
}
\label{sup_SRMHair}
\vspace{-0.4cm}
\end{figure*}

As shown in Fig.~\ref{sup_SRMHair}, we provide more reconstruction results for SRM-Hair. SRM-Hair accurately reflects the full head region and captures various hairstyles, such as afros, long hair, short hair, and ponytails. The reconstructed hair and face are separated, which enhances its potential for practical applications.

{\small
\bibliographystylesupp{ieeenat_fullname}
\bibliographysupp{hair_sup}
}

\end{document}